\PassOptionsToPackage{table}{xcolor}
\documentclass{article} 
\usepackage[final]{colm2026_conference}

\usepackage{microtype}
\usepackage{hyperref}
\usepackage{url}
\usepackage{booktabs}

\usepackage{graphicx}
\usepackage{subcaption}
\usepackage{booktabs} 

\usepackage{amsfonts}       
\usepackage{nicefrac}       
\usepackage{microtype}      
\usepackage{xcolor}         

\usepackage{xspace}
\usepackage{enumitem}
\usepackage{amsmath}
\usepackage{cleveref}

\usepackage{booktabs}
\usepackage{multirow}
\usepackage{colortbl}       
\usepackage{xcolor}         
\usepackage{graphicx}
\usepackage{wrapfig}
\usepackage{stmaryrd}
\usepackage{listings}
\usepackage{tcolorbox}
\usepackage{subcaption}
\usepackage{float}

\usepackage{lineno}

\definecolor{darkblue}{rgb}{0, 0, 0.5}
\hypersetup{colorlinks=true, citecolor=darkblue, linkcolor=darkblue, urlcolor=darkblue}

\usepackage{amssymb}
\usepackage{mathtools}
\usepackage{amsthm}

\title{SuperCoder: Assembly Program Superoptimization with Large Language Models}

\author{
\\
\textbf{Anjiang Wei\textsuperscript{1}
\quad Tarun Suresh\textsuperscript{2}
\quad Huanmi Tan\textsuperscript{3}
\quad Yinglun Xu\textsuperscript{2}}
\\
\textbf{Gagandeep Singh\textsuperscript{2}
\quad Ke Wang\textsuperscript{4}
\quad Alex Aiken\textsuperscript{1}}
\\
\textsuperscript{1}Stanford University
\quad \textsuperscript{2}University of Illinois Urbana-Champaign
\\
\textsuperscript{3}Carnegie Mellon University
\quad \textsuperscript{4}Nanjing University
}

\newcommand{\basecorrect}{61.4\%\xspace}
\newcommand{\finalcorrect}{95.0\%\xspace}

\newcommand{\finalcorrectci}{95.0 ± 0.0\%\xspace}

\newcommand{\basespeedup}{1.10×\xspace}
\newcommand{\finalspeedup}{1.46×\xspace}

\newcommand{\finalspeedupci}{1.46 ± 0.12×\xspace}

\newcommand{\numllmother}{23\xspace}

\newcommand{\name}{SuperCoder\xspace}

\newcommand{\numdataset}{8,072\xspace}

\newcommand{\claudecorrect}{51.5\%\xspace}

\newcommand{\claudespeedup}{1.43×\xspace}

\newcommand{\basemodel}{Qwen2.5-Coder-7B-Instruct\xspace}

\definecolor{lightgray}{gray}{0.9}

\lstdefinestyle{asmstyle}{
  language={[x86masm]Assembler},
  basicstyle=\ttfamily\footnotesize,
  keywordstyle=\color{blue},
  commentstyle=\color{gray},
  showstringspaces=false,
  breaklines=true,
  keepspaces=true,
  keywords={},
  morekeywords={xorl,testq,je,retq,ret,movq,movl,shrq,addq,andl,jne,popcnt,testl,jle,jmp,subl,cltd,idivl,leaq,call,decl,jg,subq},
  literate=
    {.L0:}{{\textbf{.L0:}}}1
    {.L1:}{{\textbf{.L1:}}}1
    {.L2:}{{\textbf{.L2:}}}1
    {.L3:}{{\textbf{.L3:}}}1
    {.L4:}{{\textbf{.L4:}}}1
    {.L5:}{{\textbf{.L5:}}}1
}

\definecolor{darkgreen}{RGB}{0,100,0}  

\lstdefinestyle{cstyle}{
  language=C,
  basicstyle=\ttfamily\footnotesize,
  keywordstyle=\color{blue},
  commentstyle=\color{gray},
  stringstyle=\color{red},
  showstringspaces=false,
  breaklines=false,
  keepspaces=true,
  tabsize=2,
  morekeywords={size_t,uint64_t},
  literate=
  {0}{{\textcolor{darkgreen}{0}}}1
  {1}{{\textcolor{darkgreen}{1}}}1
  {0x}{{\textcolor{darkgreen}{0x}}}2
  {ull}{{\textcolor{darkgreen}{ull}}}3
}

\begin{document}

\ifcolmsubmission
\linenumbers
\fi

\maketitle
\renewcommand\thefootnote{}
\footnotetext{Correspondence to: \texttt{anjiang@cs.stanford.edu}}
\addtocounter{footnote}{-1}

\begin{abstract}
Superoptimization is the task of transforming a program into a faster one, and ideally the very fastest possible one, while preserving its input–output behavior. In this work, we investigate whether large language models (LLMs) can serve as superoptimizers, generating assembly programs that outperform code already optimized by industry-standard compilers in end-to-end runtime. We construct the first large-scale benchmark for this problem, consisting of \numdataset assembly programs averaging 130 lines, in contrast to prior datasets restricted to 2–15 straight-line, loop-free programs. We evaluate \numllmother{} LLMs on this benchmark and find that the strongest baseline, Claude-opus-4, achieves a \claudecorrect test-passing rate and a \claudespeedup average speedup over gcc -O3. To further enhance performance, we fine-tune models with reinforcement learning, optimizing a reward function that integrates correctness and performance speedup. Starting from \basemodel{} (\basecorrect correctness, \basespeedup{} speedup), the fine-tuned model \name{} attains \finalcorrect correctness and \finalspeedup{} average speedup, with additional improvement enabled by Best-of-N sampling and iterative refinement. Our results demonstrate, for the first time, that LLMs can be applied as superoptimizers for assembly programs, establishing a foundation for future research in program performance optimization beyond compiler heuristics.
\end{abstract}

\section{Introduction}
\label{sec:intro}

Superoptimization is the task of transforming a program into a faster one, and ideally the fastest possible program, while preserving its input-output behavior. In this work, we investigate whether large language models (LLMs) can perform superoptimization by generating assembly code that surpasses the performance of compiler outputs.

Decades of research have tackled the problem of code optimization, giving rise to two main approaches. The first develops better algorithms for rule-based transformations in compilers~\citep{wolf1991loop}. However, given the vast space of possible transformations, compiler-optimized code is not guaranteed to be optimal and in practice can leave performance untapped~\citep{thomas1971catalogue,theodoridis2022finding}. The second, superoptimization, develops search algorithms that directly explore the space of all possible programs, aiming to discover the correct variant with the best performance rather than relying on a fixed set of transformation rules~\citep{schkufza2013stochastic}.

Recent learning-based compiler optimization methods operate within the structure of a compiler, for example, by selecting optimization pass sequences or compiler flags~\citep{cummins2025llm,pancompiler}. While effective, these approaches are constrained to expert-defined transformations. Superoptimization is a more aggressive form of program optimization that directly rewrites code without being restricted to predefined rules, enabling a much larger exploration space. By operating at the assembly level, superoptimization can discover novel optimizations that go beyond what compilers can achieve.  The existing superoptimization literature has focused on very short, straight-line assembly programs without loops. Prior work has primarily relied on search heuristics that do not scale to larger programs~\citep{schkufza2013stochastic,phothilimthana2016scaling,koenig2021adaptive}, and prior datasets include programs of at most 15 lines of assembly~\citep{artifact}.

\begin{figure}[!tb]
    \centering
    \includegraphics[width=0.9\linewidth]{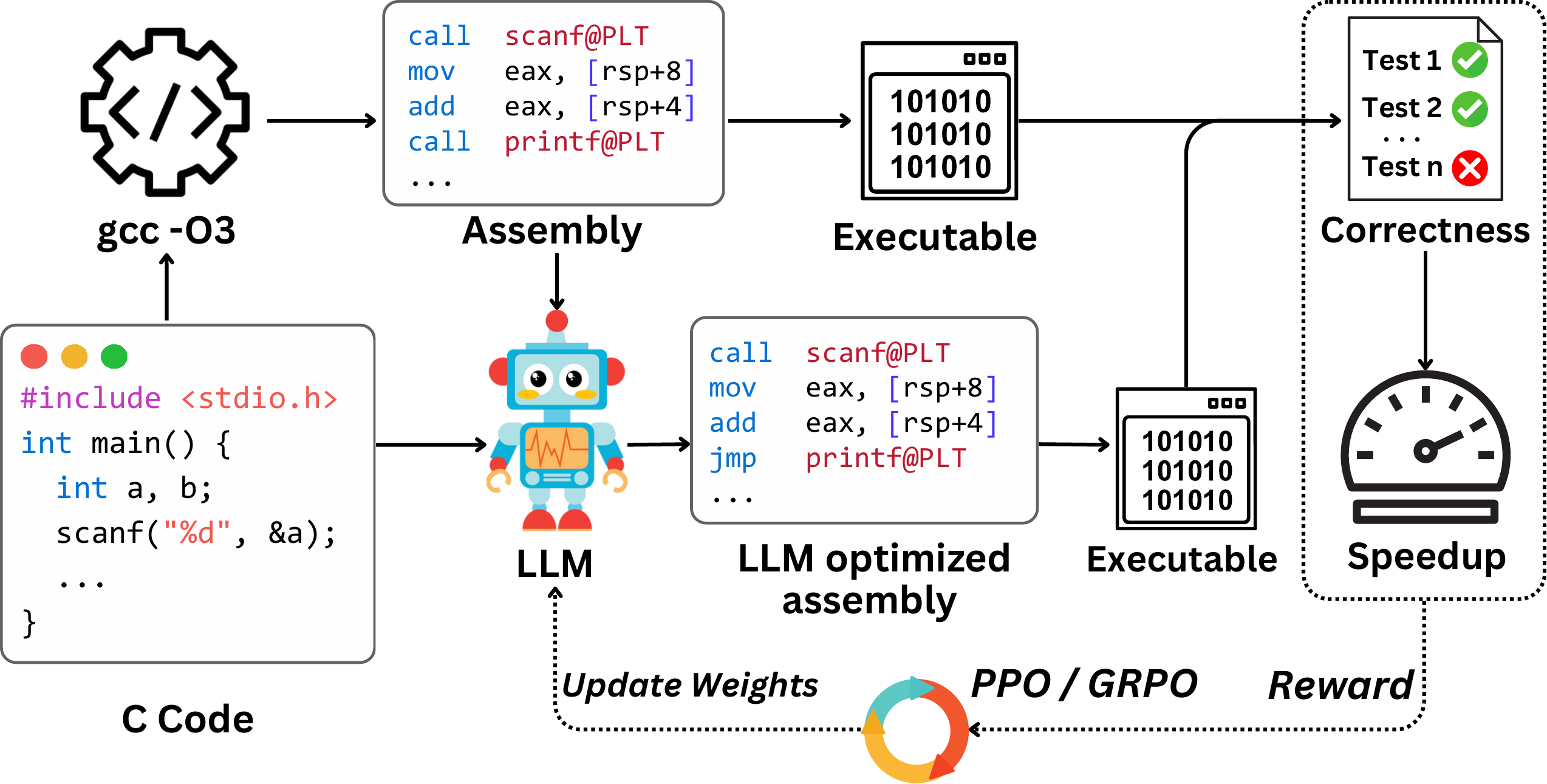}
    \caption{
Overview of the assembly code optimization task. Given a C program and its baseline assembly from gcc -O3, an LLM is fine-tuned with PPO or GRPO to generate improved assembly. The reward function reflects correctness and performance based on test execution.
}
    \label{fig:intro}
\end{figure}

In this work, we explore the use of LLMs as a superoptimizer to improve the performance of assembly code. In contrast to most prior work on code generation from natural language~\citep{chen2021evaluating,austin2021program,zhuo2024bigcodebench}, we tackle a fundamentally different and more technically demanding task: improving assembly code that has already been optimized by the industry-standard compiler at its highest optimization level (gcc -O3). Production compilers have been refined over decades of expert-driven development, and surpassing them remains a challenge.

Unlike high-level programming languages (e.g., Python or C), large-scale, high-quality assembly datasets are scarce. As the first study in this direction, we construct a dataset of \numdataset assembly programs. Each instance includes input–output test cases and baseline assembly generated by the compiler at its highest optimization level (gcc -O3), which serves as the starting point for further optimization. In contrast, the datasets commonly used in the superoptimization community~\citep{warren2013hacker,schkufza2013stochastic,artifact} are either limited in size or consist of short examples with 2 to 15 instructions without loops. Our dataset is substantially larger, with assembly programs averaging 130 lines and including loops. Our test suites achieve 96.2\% line and 87.3\% branch coverage, demonstrating strong test quality.

Beyond evaluating existing models, we apply reinforcement learning (RL) to further enhance their capabilities. We employ widely adopted algorithms, including Proximal Policy Optimization (PPO) and Group Relative Policy Optimization (GRPO), to train an LLM with a reward function that integrates both correctness and performance speedup. Prior work on LLM-based performance optimization has explored alternative methodologies such as supervised fine-tuning~\citep{shypula2023learning}, chain-of-thought prompting~\citep{liu2024evaluating}, agent-based frameworks~\citep{wei2025improving,wei2025astra}, and preference learning~\citep{du2024mercury}. We optimize speedup explicitly in the reward function, making RL well-suited to superoptimization.

To our knowledge, SuperCoder is among the first works to use reward-based RL for LLM-based code performance optimization that directly optimizes end-to-end runtime speedup under correctness constraints. Prior learning-based compiler optimization methods instead optimize compiler-side surrogates such as IR instruction count or code size~\citep{deng2025compilerdream,cummins2025llm}, rather than end-to-end runtime. In contrast, SuperCoder formulates the task as post-compiler assembly superoptimization, providing a more direct path to real-world performance improvements.

We evaluate \numllmother LLMs on this task and find that the best-performing model, Claude-opus-4, achieves a \claudecorrect test-passing rate and an average speedup of \claudespeedup over the compiler-optimized baseline (gcc -O3). Our reinforcement learning approach is highly effective: starting from the base model \basemodel, which achieves \basecorrect correctness and a modest \basespeedup{} speedup, the fine-tuned model \name attains \finalcorrect correctness and \finalspeedup{} average speedup, with further improvement enabled by Best-of-N sampling and iterative refinement.

In summary, our contributions are as follows:

\begin{itemize}
\item We are the first to introduce superoptimization as a task for LLMs, a technically demanding challenge that aims to improve the end-to-end runtime of the compiler-optimized assembly code.

\item We construct the first large-scale dataset of \numdataset assembly programs, with examples averaging 130 lines of code. 

\item We evaluate \numllmother{} LLMs on the benchmark and show that RL-based training substantially improves performance: fine-tuning \basemodel{} (\basecorrect correctness, \basespeedup{} speedup) results in \name{} with \finalcorrect correctness and \finalspeedup{} speedup, with further gains enabled by Best-of-N sampling and iterative refinement.
\end{itemize}

\section{Related Work}
\label{sec:related}

\paragraph{Large Language Models for Code.}
Large language models (LLMs) have been extensively evaluated on code generation benchmarks such as HumanEval~\citep{chen2021evaluating}, MBPP~\citep{austin2021program}, APPS~\citep{hendrycks2021measuring}, and others~\citep{evalplus,li2024evocodebench,xia2024top,zhuo2024bigcodebench}. Numerous models have been developed for code generation, including Codex~\citep{chen2021evaluating}, AlphaCode~\citep{li2022competition}, StarCoder~\citep{li2023starcoder}, DeepSeek-Coder~\citep{guo2024deepseek}, and Code Llama~\citep{roziere2023code}. LLMs have also been applied to software engineering tasks such as program repair~\citep{xia2022less,xia2023automated}, testing~\citep{xia2023universal,deng2024large}, transpilation~\citep{yang2024exploring,bhatia2024verified}, and synthesis~\citep{wei2025codearc}. SWE-bench~\citep{jimenez2023swe} provides a benchmark for resolving real GitHub issues, with agent-based approaches~\citep{yang2024swe,xia2024agentless,wei2025swe} further improving issue resolution.

Recent work has explored LLMs for improving program performance, including automatic parallelization~\citep{tehranijamsaz2024coderosetta}, Python optimization~\citep{du2024mercury,liu2024evaluating,huang2024effilearner}, C++ performance~\citep{shypula2023learning}, and low-level code optimization~\citep{wei2025improving,ouyang2025kernelbench}. Prior learning-based compiler optimization methods operate within the structure of a compiler, selecting optimization pass sequences~\citep{cummins2024meta,deng2025compilerdream} or compiler flags~\citep{cummins2025llm}, and typically optimize compiler-side surrogates such as IR instruction count~\citep{cummins2024meta} or code size~\citep{deng2025compilerdream,cummins2025llm}, rather than end-to-end runtime. LLM-Vectorizer~\citep{taneja2025llm} offers a formally verified solution for auto-vectorization, a specific compiler pass. In contrast, we formulate the task as post-compiler assembly superoptimization: starting from gcc \texttt{-O3} output, it directly rewrites assembly and is trained to optimize measured execution-time speedup, providing a more direct path to real-world performance improvements. Our method is not intended to replace safer compiler-surrogate approaches. Instead, it is a complementary route for exploring optimizations beyond the compiler-defined search space.

\paragraph{Learning-Based Code Optimization.}
The space of code optimization is vast, and many approaches have leveraged machine learning to improve program performance. A classic challenge in compilers is the phase-ordering problem, where performance depends heavily on the sequence of optimization passes. AutoPhase~\citep{haj2020autophase} uses deep reinforcement learning to tackle this, while Coreset~\citep{liang2023learning} employs graph neural networks (GNNs) to guide optimization decisions. Modern compilers apply extensive rewrite rules but offer no guarantee of optimality. Superoptimization seeks the most efficient program among all semantically equivalent variants of the compiler output. Traditional methods use stochastic search, such as MCMC sampling~\citep{schkufza2013stochastic}, with follow-up work improving scalability~\citep{phothilimthana2016scaling,bunel2016learning} and extending to broader domains~\citep{sharma2015conditionally,churchill2017sound}. These rely on formal verification for correctness, restricting them to date to small, loop-free programs. In contrast, our approach uses test-based validation, enabling optimization of general programs with loops. With the rise of deep learning, substantial attention has turned to optimizing GPU kernel code. AutoTVM~\citep{chen2018learning} pioneered statistical cost model-based search for CUDA code optimization, followed by methods such as Ansor~\citep{zheng2020ansor}, AMOS~\citep{zheng2022amos}, and others~\citep{shao2022tensor,zhao2024felix,wu2024mirage}.

More recently, LLMs have been explored as code optimizers~\citep{shypula2023learning,grubisic2024compiler,wei2025improving,wei2025astra}, with reinforcement learning guiding generation through rewards based on unit-test correctness~\citep{le2022coderl,shojaee2023execution,liu2023rltf,dou2024stepcoder,wei2025swe} or preference signals~\citep{liu2024learning,du2024mercury}. To our knowledge, this is among the first works to apply RL to optimize end-to-end code performance with LLMs, with concurrent efforts on CUDA kernel optimization~\citep{li2025cuda,baronio2025kevin}.

\section{Methodology}
\label{sec:method}

\subsection{Task Definition}
\label{subsec:task}

Let \( C \) be a program written in a high-level language such as C. A modern compiler like gcc can compile \( C \) into an x86-64 assembly program \( P = gcc(C) \), which can then be further assembled into an executable binary. The assembly program \( P \) serves as an intermediate representation that exposes low-level optimization opportunities, making it suitable for aggressive performance improvement. We assume the semantics-preserving nature of the compilation process, i.e., \( \llbracket C \rrbracket = \llbracket P \rrbracket \), so that the behavior of the assembly program \( P \) is identical to that of the source program \( C \).

In theory, the goal is to produce a program \( P' \) that is functionally equivalent to \( P \) across the entire input space \( \mathcal{X} \), i.e., \( P(x) = P'(x) \) for all \( x \in \mathcal{X} \). Since verifying this property is undecidable in general, we approximate equivalence using a finite test set \( \mathcal{T} = \{(x_i, y_i)\}_{i=1}^n \), where each input-output pair \( (x_i, y_i) \) captures the expected behavior of \( C \).

We say that an assembly program \( P' \) is \emph{valid} if it can be successfully assembled and linked into an executable binary. Let \( \texttt{valid}(P') \in \{\texttt{True}, \texttt{False}\} \) denote this property. We therefore define the set of \emph{correct} programs as:
\[
\mathcal{S}(P) = \left\{ P' \mid \texttt{valid}(P') \; \land \; \forall (x_i, y_i) \in \mathcal{T},\; P'(x_i) = y_i \right\}.
\]

\paragraph{Performance and Speedup.}
Let \( t(P) \) denote the execution time of \( P \) on the test set \( \mathcal{T} \), and let \( t(P') \) be the corresponding execution time for \( P' \). The speedup of \( P' \) relative to \( P \) is defined as follows. If the LLM-generated program is invalid or slower, we fall back to the baseline and assign a speedup of 1.

\[
\text{Speedup}(P') =
\begin{cases}
\frac{t(P)}{t(P')} & \text{if } P' \in \mathcal{S}(P) \text{ and } t(P') < t(P), \\
1 & \text{otherwise}.
\end{cases}
\]

\paragraph{Optimization Objective.}
The objective is to generate a candidate program \( P' \) that maximizes \( \text{Speedup}(P') \). Only programs in \( \mathcal{S}(P) \) are eligible for speedup; any candidate that fails to compile into a binary or produces incorrect outputs is assigned a default speedup of 1. This reflects a practical fallback: when the generated program is invalid, the system can revert to the baseline \( P \), compiled with gcc -O3, which defines the 1× reference performance. Although \( \mathcal{S}(P) \) captures the correctness criteria, we do not restrict the LLM to generate only valid programs. Instead, the model produces arbitrary assembly code, and correctness is validated post hoc via compilation and test execution. We train an LLM using reinforcement learning (see \Cref{subsec:rl}) to generate candidates that both satisfy correctness and achieve performance improvements.

\subsection{Dataset Construction}
\label{subsec:dataset}

We construct our dataset using C programs from CodeNet~\citep{puri2021codenet}, a large-scale corpus of competitive programming submissions. CodeNet is a well-established and widely used benchmark in the AI-for-code community~\citep{li2022competition,shypula2023learning}. Each dataset instance is a tuple \( (C, P, \mathcal{T}) \), where \( C \) is the original C source code, \( P = gcc(C) \) is the corresponding x86-64 assembly generated by compiling \( C \) with gcc at the -O3 optimization level, and \( \mathcal{T} = \{(x_i, y_i)\}_{i=1}^n \) is the test set. Since not all CodeNet problems include test inputs, we adopt those provided by prior work~\citep{li2022competition} to define \( x_i \), but discard their output labels. Instead, we regenerate each output \( y_i \) by executing the original submission on input \( x_i \), as many CodeNet programs are not accepted solutions, and even accepted ones do not reliably pass all test cases. Given the scale of CodeNet, which contains over 8 million C and C++ submissions, we sample a subset for this study.

\subsection{Reinforcement Learning}
\label{subsec:rl}
We conceptualize our task as a standard contextual multi-armed bandit problem~\citep{lu2010contextual}, defined by a context space \( \mathcal{S} \), an action space \( \mathcal{A} \), and a reward function \( r : \mathcal{S} \times \mathcal{A} \rightarrow \mathbb{R} \). Each context \( s \in \mathcal{S} \) represents a problem instance, comprising the source program \( C \), its baseline assembly \( P \), and the associated test cases \( \mathcal{T} \). An action \( a \in \mathcal{A} \) corresponds to generating a candidate assembly program \( \tilde{P} \). The reward function \( r(s, a) \) evaluates the quality of the generated program based on correctness and performance. We will describe different designs of the reward function later. A policy \( \pi : \mathcal{S} \rightarrow \Delta(\mathcal{A}) \) maps a context \( s \) to a probability distribution over actions and samples an action \( a \in \mathcal{A} \) stochastically. Given a distribution \( \mu \) over problem instances, the expected performance of a policy \( \pi \) under reward function \( r \) is expressed as \( \mathbb{E}_{s \sim \mu, a \sim \pi(\cdot \mid s)} \left[ r(s, a) \right] \). The objective is to find a policy that maximizes this expected reward.

\paragraph{Optimization with PPO and GRPO.}
We train the policy using two policy-gradient algorithms: \emph{Proximal Policy Optimization} (PPO)~\citep{schulman2017proximal} and \emph{Group Relative Policy Optimization} (GRPO)~\citep{shao2024deepseekmath}. PPO stabilizes training by constraining each update to remain close to the previous policy. It maximizes a clipped surrogate objective of the form
\(
\mathbb{E}_{s, a} \left[ \min\left( \rho(\theta) \hat{A},\; \operatorname{clip}(\rho(\theta), 1 - \epsilon, 1 + \epsilon)\, \hat{A} \right) \right],
\)
where \( \rho(\theta) = \pi_\theta(a \mid s) / \pi_{\theta_{\text{old}}}(a \mid s) \), \( \hat{A} \) is the estimated advantage, and \( \epsilon \) controls the clipping range. GRPO, in contrast, compares rewards among a group of sampled outputs and assigns a higher likelihood to relatively stronger ones. In both algorithms, rewards are based on the correctness and execution time of the generated program.

\paragraph{Reward Function Design.}
As defined in our contextual bandit setup, the reward function \( r : \mathcal{S} \times \mathcal{A} \rightarrow \mathbb{R} \) assigns a scalar score to each (context, action) pair. Each context \( s \in \mathcal{S} \) consists of the source program \( C \), the baseline assembly \( P \), and a test set \( \mathcal{T} = \{(x_i, y_i)\}_{i=1}^n \). An action \( a \in \mathcal{A} \) corresponds to a generation procedure that produces a candidate assembly program \( \tilde{P} = \texttt{gen}(a) \).

We define two auxiliary metrics for computing reward:
\[
\operatorname{pass}(s, a) = \tfrac{1}{|\mathcal{T}|} \sum_{(x, y) \in \mathcal{T}} \mathbf{1}[\tilde{P}(x) = y],
\]
\[
\operatorname{speedup}(s, a) = t(P) / t(\tilde{P}),
\]
which denote the fraction of test cases passed and the speedup of the generated program \( \tilde{P} \) relative to the baseline \( P \). We use the following reward function during training:

\[
r(s, a) =
\begin{cases}
0, & \text{if } \operatorname{pass}(s, a) < 1, \\[4pt]
\operatorname{speedup}(s, a), & \text{if } \operatorname{pass}(s, a) = 1.
\end{cases}
\]

If a generated program fails to compile or does not pass all tests, its reward is set to 0, with no partial credit for partial correctness. Only when the code compiles and passes all tests is a positive reward assigned, equal to the achieved speedup.

\subsection{Best-of-N Sampling, Supervised Fine-Tuning, and Iterative Refinement}
\label{subsec:sft}

\paragraph{Best-of-N Sampling.}
Generating multiple candidate programs and selecting the strongest one is a well-established strategy for improving code generation quality~\citep{li2022competition,ehrlich2025codemonkeys}. In our setting, the best candidate refers to the program that is correct while achieving the fastest execution time. Best-of-N sampling is an inference-time technique that can boost performance, but it incurs additional cost.

\paragraph{Supervised Fine-Tuning.} To obtain training targets for supervised fine-tuning, we require reference solutions. However, superoptimization is inherently open-ended: beyond the compiler baseline, there is no unique ground-truth program, and multiple distinct solutions may exist. We apply best-of-8 sampling with the base model and treat the highest-quality candidate as the ground truth. We then fine-tune the model using LoRA~\citep{hu2022lora}.

\paragraph{Iterative Refinement.} Iterative refinement is a complementary inference-time technique that can be applied to any model to further improve its outputs. After each trial, we feed back the model’s previous attempt. If the generated program fails to compile or fails any test cases, we include the corresponding compiler errors or test failures in the next prompt.
\section{Experimental Setup}
\label{sec:setup}

\paragraph{Dataset.}

The final dataset contains 7{,}872 training programs and 200 evaluation programs, with average program lengths and test case counts summarized in Appendix~\ref{subsec:datasetstats}. Rather than relying directly on the original input-output pairs, we re-run each program on its inputs to generate correct outputs, thereby fixing errors in prior datasets. The resulting test suites of our evaluation dataset achieve an average of 96.2\% line coverage and 87.3\% branch coverage, demonstrating high test quality. We also make sure that test cases are withheld from the model.

\paragraph{Prompts.} For each instance, we construct a prompt that includes the original C program along with the generated assembly using gcc -O3. Test cases are withheld from the model. We show the prompt template in Appendix~\ref{subsec:prompttemplate}.

\paragraph{Models.} We evaluate \numllmother{} state-of-the-art language models spanning a diverse range of architectures. Our benchmark includes frontier proprietary models such as gpt-4o~\citep{achiam2023gpt}, o4-mini, gemini-2.0-flash-001~\citep{team2023gemini}, and claude-3.7-sonnet, as well as open-source families such as Llama~\citep{touvron2023llama}, DeepSeek~\citep{liu2024deepseek}, and Qwen~\citep{hui2024qwen2}. In addition, we include models distilled from DeepSeek-R1~\citep{guo2025deepseek} based on Qwen and Llama. Finally, we evaluate recent compiler-oriented foundation models~\citep{cummins2024meta,cummins2025llm}, pretrained on assembly code and derived from Code Llama, with a design focus on compiler tasks (listed as llm-compiler in \Cref{tab:main}). All open-source models are instruction-tuned.

\paragraph{Implementation.}
We implement our customized reinforcement learning reward functions within the \textsc{verl} framework~\citep{sheng2024hybridflow}, which enables fine-tuning of LLMs using PPO and GRPO. As part of this setup, we build a task-specific environment that handles program compilation, test execution, and runtime measurement, as detailed in \Cref{subsec:rl}. This environment provides the model with direct scalar feedback based on both functional correctness and execution performance.

\paragraph{Metrics.}

We evaluate each model using both correctness and performance metrics. \emph{Compile pass} is the percentage of problems for which the generated assembly compiles to a binary executable successfully, while \emph{test pass} is the percentage of problems where the compiled code passes all test cases. For performance, we measure the relative speedup over the gcc -O3 baseline. We report the \emph{25th}, \emph{50th} (median), and \emph{75th} percentiles of speedup, along with the \emph{average speedup} over the entire evaluation set.

\paragraph{Performance Measurement.}
To ensure an accurate performance evaluation, we use hyperfine~\citep{hyperfine}, a benchmarking tool that reduces measurement noise by performing warmup runs followed by repeated timed executions. For each program's execution, we discard the first three runs and report the average runtime over the next ten runs.

We provide additional details of the experimental setup in \Cref{subsec:app:setup}.

\section{Results}
\label{sec:result}

\subsection{Evaluation of Different Models}

\begin{table*}[!tb]
\small
\centering
\begin{tabular}{@{}lrrrrrrr}
\toprule
\multirow{2}{*}{\textbf{Model}} & 
\multirow{2}{*}{\shortstack{\textbf{Compile}\\\textbf{Pass}}} & 
\multirow{2}{*} {\shortstack{\textbf{Test}\\\textbf{Pass}}} & 
\multicolumn{3}{c}{\textbf{Speedup Percentiles}} & 
\multirow{2}{*}{\shortstack{\textbf{Average}\\\textbf{Speedup}}} \\
& & & 25th & 50th & 75th & \\
\midrule
DS-R1-Distill-Qwen-1.5B & 0.0\% & 0.0\% & 1.00× & 1.00× & 1.00× & \cellcolor{lightgray}1.00× \\
DeepSeek-R1 & 0.0\% & 0.0\% & 1.00× & 1.00× & 1.00× & \cellcolor{lightgray}1.00× \\
DS-R1-Distill-Llama-70B & 5.5\% & 0.0\% & 1.00× & 1.00× & 1.00× & \cellcolor{lightgray}1.00× \\
DS-R1-Distill-Qwen-14B & 11.5\% & 0.5\% & 1.00× & 1.00× & 1.00× & \cellcolor{lightgray}1.00× \\
gpt-4o-mini & 44.5\% & 1.0\% & 1.00× & 1.00× & 1.00× & \cellcolor{lightgray}1.00× \\
Llama-4-Maverick-17B & 77.5\% & 7.0\% & 1.00× & 1.00× & 1.00× & \cellcolor{lightgray}1.02× \\
Llama-3.2-11B & 84.0\% & 21.0\% & 1.00× & 1.00× & 1.00× & \cellcolor{lightgray}1.02× \\
gpt-4o & 81.0\% & 5.0\% & 1.00× & 1.00× & 1.00× & \cellcolor{lightgray}1.02× \\
Llama-4-Scout-17B & 68.5\% & 5.5\% & 1.00× & 1.00× & 1.00× & \cellcolor{lightgray}1.02× \\
o4-mini & 25.0\% & 4.5\% & 1.00× & 1.00× & 1.00× & \cellcolor{lightgray}1.02× \\
gemini-2.0-flash-001 & 57.5\% & 4.0\% & 1.00× & 1.00× & 1.00× & \cellcolor{lightgray}1.03× \\
Qwen2.5-72B & 59.5\% & 7.5\% & 1.00× & 1.00× & 1.00× & \cellcolor{lightgray}1.03× \\
Llama-3.2-90B & 82.5\% & 15.0\% & 1.00× & 1.00× & 1.00× & \cellcolor{lightgray}1.05× \\
Qwen2.5-Coder-7B & 77.9\% & 61.4\% & 1.00× & 1.00× & 1.00× & \cellcolor{lightgray}1.10× \\
gpt-5 & 78.5\% & 6.0\% & 1.00× & 1.00× & 1.00× & \cellcolor{lightgray}1.13× \\
DeepSeek-V3 & 94.0\% & 43.0\% & 1.00× & 1.00× & 1.40× & \cellcolor{lightgray}1.21× \\
claude-3.7-sonnet & 94.5\% & 58.5\% & 1.00× & 1.10× & 1.45× & \cellcolor{lightgray}1.22× \\
claude-sonnet-4 & 87.0\% & 37.0\% & 1.00× & 1.00× & 1.95× & \cellcolor{lightgray}1.30× \\
claude-opus-4 & 90.0\% & 51.5\% & 1.00× & 1.58× & 2.03× & \cellcolor{lightgray}1.43× \\
\midrule
llm-compiler-7b-ftd & 2.0\% & 2.0\% & 1.00× & 1.00× & 1.00× & \cellcolor{lightgray}1.00× \\
llm-compiler-13b-ftd & 2.5\% & 2.0\% & 1.00× & 1.00× & 1.00× & \cellcolor{lightgray}1.01× \\
llm-compiler-7b & 55.0\% & 54.0\% & 1.00× & 1.00× & 1.00× & \cellcolor{lightgray}1.09× \\
llm-compiler-13b & 60.5\% & 59.5\% & 1.00× & 1.27× & 1.63× & \cellcolor{lightgray}1.34× \\
\bottomrule
\end{tabular}
\vspace{1em}
\caption{Comparison of LLMs on our assembly optimization benchmark. We report compilation success rate, test pass rate, and average speedup over the gcc -O3 baseline. Speedup is averaged across all test inputs, with each input evaluated over ten runs.}
\label{tab:main}
\end{table*}

\paragraph{Main Results.} \Cref{tab:main} reports results across evaluated models. Most perform poorly on this task, with only a few demonstrating effectiveness as superoptimizers. Most models struggle to generate performant assembly: the majority yield only 1.00× speedup, with low compile and test pass rates. Among all models, claude-opus-4 and claude-sonnet-4 perform best, achieving average speedups of 1.43× and 1.30×, respectively. Compiler foundation models (prefixed with llm-compiler-)~\citep{cummins2024meta,cummins2025llm}, are pretrained on assembly code and derived from Code Llama, with a design focus on compiler tasks. Among them, llm-compiler-13b achieves a notable 1.34× speedup, whereas the fine-tuned variants (-ftd) perform poorly, likely because they were adapted for different tasks such as reducing code size and compiler flag tuning. These results suggest that while superoptimization is inherently difficult, some LLMs can be effective superoptimizers. This also indicates that models for other compiler-related tasks, such as compiler flag tuning, may not be effective in directly optimizing assembly performance.

\paragraph{Failure Modes.} Models that are expected to achieve strong results (e.g., DeepSeek-R1, GPT-4o) perform poorly on the task of superoptimization, motivating our analysis of their failure modes. We find that DeepSeek-R1 consistently fails to generate valid assembly code, resulting in a 0\% compilation rate. DeepSeek-R1 often produces verbose analysis instead of executable code, spending its entire output length on reasoning about instruction semantics and potential optimizations without actually implementing them.

We further analyze the failure modes of GPT-4o, which achieves a high compilation rate (81.0\%) but exhibits poor correctness (only 5.0\% test pass rate). The primary correctness issues are as follows: (1) missing critical directives, such as stack canary initialization and \texttt{.cfi\_*} metadata, which often lead to runtime crashes; (2) incorrect function call conventions, where repeated system calls like \texttt{scanf} are made without proper argument setup, causing undefined behavior; (3) semantic errors in core computations, including incorrect pointer usage or altered algorithm logic, which produce wrong outputs even when the code runs; and (4) over-simplified stack or register management, resulting in memory errors or invalid control flow. In summary, GPT-4o generates syntactically valid assembly but frequently violates low-level conventions necessary for correct and reliable execution.

\subsection{Effectiveness of RL Training}
\label{subsec:rlresult}

\textbf{Improvement.} \Cref{tab:supercoder} presents the results of RL training, averaged over 5 runs with 95\% confidence intervals to provide more statistical confidence in the reported improvements. We select \basemodel for RL training due to its strong test pass rate (\basecorrect) among models. After RL training with PPO, the fine-tuned model \name{} attains \finalcorrect correctness and improves average speedup from \basespeedup{} to \finalspeedup{}. Its speedup percentiles are 1.17 ± 0.03× (25th), 1.35 ± 0.04× (50th), and 1.64 ± 0.08× (75th) respectively, outperforming the majority of evaluated models.

\textbf{PPO versus GRPO.} We evaluate both PPO-trained and GRPO-trained models and find their performance to be nearly identical. \name{} trained with GRPO attains 94.7 ± 0.6\% correctness and 1.44 ± 0.07× average speedup, which is comparable to \name{} trained with PPO (\finalcorrectci correctness and \finalspeedupci average speedup).

\begin{table*}[!tb]
\small
\centering
\begin{tabular}{@{}lccc}
\toprule
\textbf{Model} & 
\textbf{Compile Pass} & 
\textbf{Test Pass} & 
\textbf{Average Speedup} \\
\midrule
Qwen2.5-Coder-7B (Base) & 77.9 $\pm$ 0.8\% & 61.4 $\pm$ 0.5\% & \cellcolor{lightgray}1.10 $\pm$ 0.01$\times$ \\
\name (GRPO) & 95.0 $\pm$ 0.0\% & 94.7 $\pm$ 0.6\% & \cellcolor{lightgray}1.44 $\pm$ 0.07$\times$ \\
\name (PPO) & 96.0 $\pm$ 0.0\% & 95.0 $\pm$ 0.0\% & \cellcolor{lightgray}1.46 $\pm$ 0.12$\times$ \\
\name (Supervised fine-tuning) & 95.5 $\pm$ 0.0\% & 92.5 $\pm$ 0.0\% & \cellcolor{lightgray}1.39 $\pm$ 0.05$\times$ \\
\bottomrule
\end{tabular}
\vspace{1em}
\caption{Performance of the base model and the models trained with RL or supervised fine-tuning. Results include compilation success, test pass rates, and average speedup, reported over 5 runs with 95\% confidence intervals.}
\label{tab:supercoder}
\end{table*}

\subsection{Results from Supervised Fine-Tuning and Inference-Time Methods}
\label{subsec:sftresult}

\paragraph{Supervised Fine-Tuning.} We describe our supervised fine-tuning approach in \Cref{subsec:sft}. \Cref{tab:main} reports results averaged over five runs with 95\% confidence intervals. While supervised fine-tuning improves performance, RL achieves slightly stronger results. We believe that RL is a natural fit for the open-ended nature of superoptimization, as RL directly optimizes for correctness and speedup rather than imitating existing examples.

\begin{figure}[!tb]
    \centering
    \begin{minipage}[t]{0.45\textwidth}
        \centering
        \includegraphics[width=\linewidth]{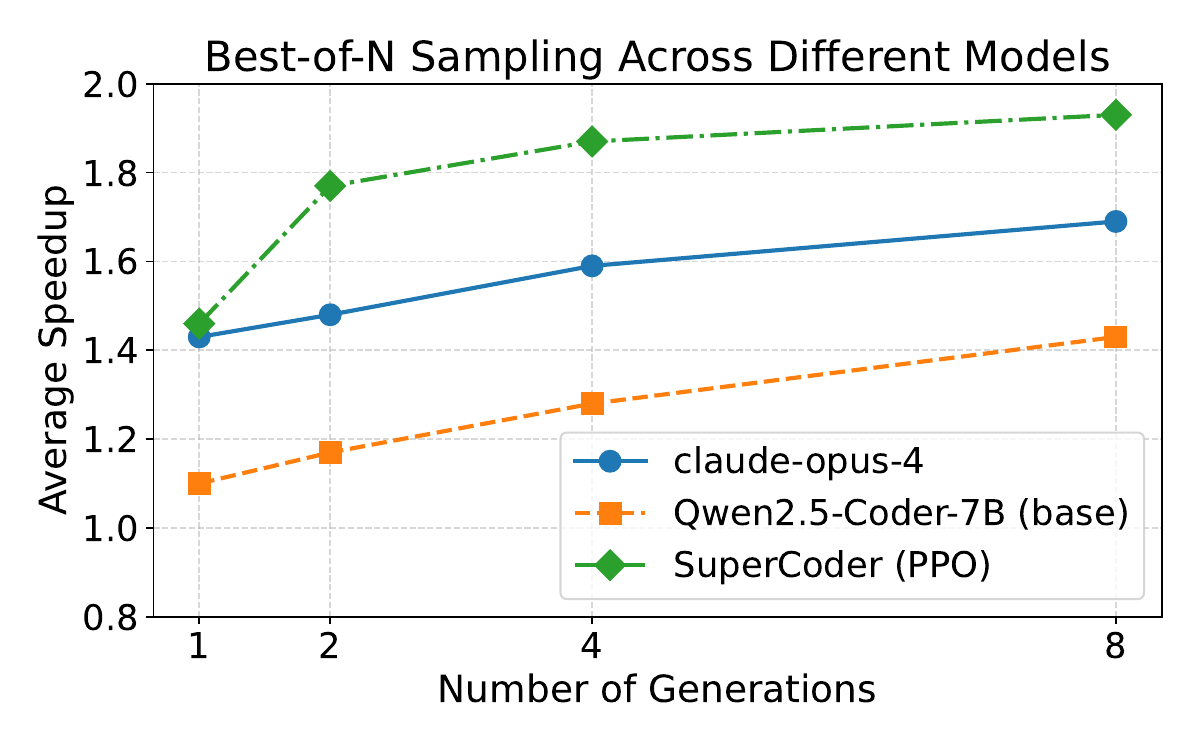}
        \vspace{-2em}
        \caption{Best-of-N sampling results.}
        \label{fig:best}
    \end{minipage}
    \hfill
    \begin{minipage}[t]{0.45\textwidth}
        \centering
        \includegraphics[width=\linewidth]{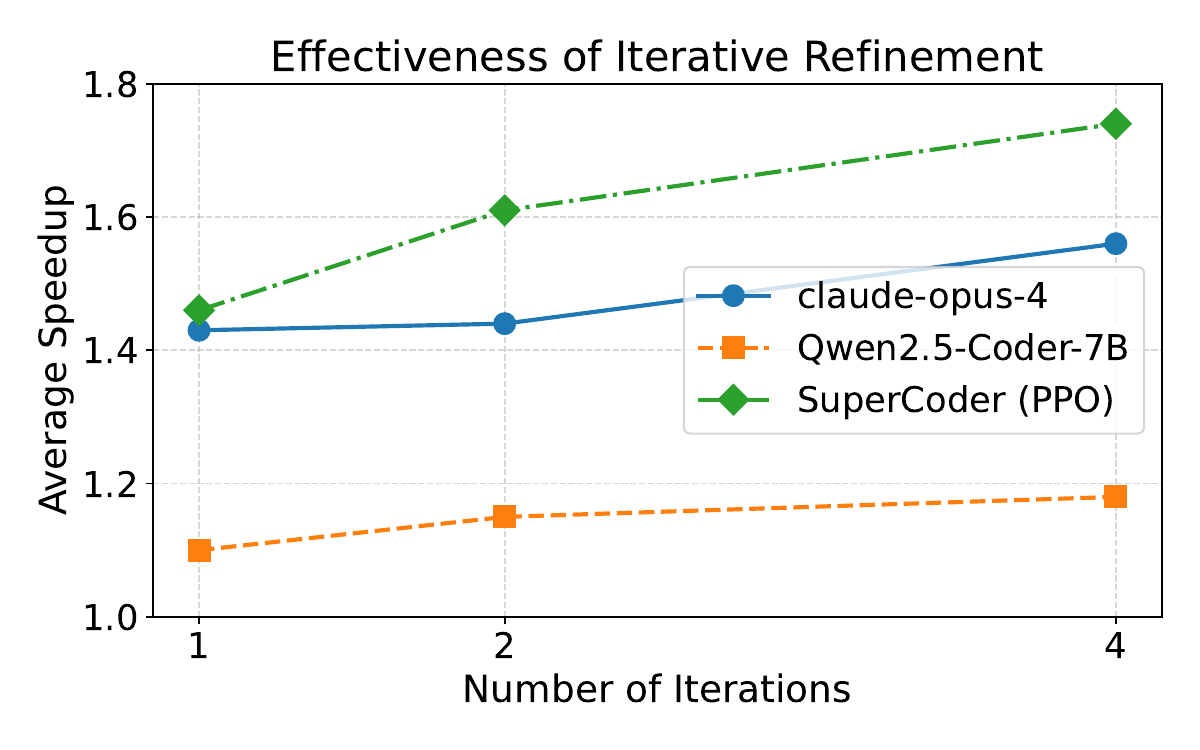}
        \vspace{-2em}
        \caption{Iterative refinement results.}
        \label{fig:iterative}
    \end{minipage}
\vspace{-1.5em}
\end{figure}

\paragraph{Best-of-N Sampling.} We evaluate best-of-N sampling for three models: claude-opus-4 (the strongest baseline in \Cref{tab:main}), Qwen2.5-Coder-7B (base), and SuperCoder (our PPO-trained model). Results are shown in \Cref{fig:best}. Notably, the base model’s best-of-8 speedup is close to the PPO-trained model’s best-of-1 result, and the RL-trained model itself can still be improved with best-of-N sampling (i.e., from 1.46× in the single-sample setting to 1.93× with best-of-8 sampling).

\paragraph{Iterative Refinement.} \Cref{fig:iterative} shows the results of iterative refinement, where the model receives feedback about compilation failures, test failures, or performance for self-reflection. All three models exhibit improvements as the number of refinement iterations increases, with the effect being most pronounced for our RL fine-tuned model.

Although Best-of-N sampling performs better under the default budget, iterative refinement can surpass it when given more iterations. Extending iterative refinement to six iterations increases claude-opus-4's mean speedup to 1.68×, slightly exceeding the 1.59× achieved by best-of-4 sampling. This suggests that iterative refinement can continue to benefit from additional feedback, although Best-of-N sampling generally appears more inference-efficient because it explores a more diverse set of candidates in parallel.

\subsection{Analysis of Program Transformations}
\label{subsec:analysis}

To better understand why LLMs can further optimize assembly programs already optimized by industry-standard compilers, we analyze all 200 evaluation programs by comparing the gcc -O3 output with the assembly generated by claude-opus-4 with best-of-8 sampling. With the assistance of LLMs and human validation, we categorize the program transformations into nine types: 1) loop restructuring; 2) instruction selection; 3) algorithmic simplification, which replaces complex or custom logic with simpler algorithms or standard library routines; 4) stack canary removal; 5) register allocation; 6) branch elimination; 7) address calculation optimization; 8) function call optimization; and 9) arithmetic optimization. Appendix~\ref{subsec:app:trans} presents a detailed category-wise description and analysis, reporting the frequency of each transformation category among all transformed programs that pass all tests. Because performance optimization typically involves multiple interrelated decisions, it is difficult to automatically isolate which types of program transformations are responsible for observed performance improvements.
\section{Discussion}
\label{sec:discussion}

\paragraph{Alternative Reward Function Design.}
We evaluate an alternative reward design that gives $-1$ for compilation failure, partial credit for passing some tests, and $1 + \alpha \cdot \text{speedup}$ once all tests pass. This yields an average speedup of 1.38×, and varying $\alpha$ (5 or 10) has little effect. The result remains below \finalspeedup{}, suggesting that directly optimizing the terminal speedup reward is more effective.

\paragraph{Few-shot Prompting.}
We evaluate 0-shot, 2-shot, and 4-shot prompting across three models. \Cref{tab:shots} shows that adding more in-context examples does not reliably improve performance and often degrades it. This observation is consistent with prior work~\citep{shypula2023learning}, which finds that few-shot examples can bias the model and lead to incorrect task understanding.

\paragraph{Prompt Evolution Framework.} We experimented with GEPA~\citep{agrawal2025gepa}, an evolutionary prompting framework that uses natural language reflection to learn high-level optimization rules from trial and error. We used gpt-4o as the model under evaluation and gpt-5 as the reflection model. GEPA yielded only modest gains: compilation pass increased from 81.0\% to 84.0\% and test pass from 5.0\% to 7.5\%, while performance speedup remained essentially unchanged. We suspect this is because assembly optimization requires substantial domain knowledge that is difficult to capture by modifying the prompt alone.

\paragraph{Direct Compilation from C.}
We also examine a more challenging setting where gcc -O3 assembly is not provided. Instead, LLMs receive only the C source code and are asked to generate assembly directly. This setup leads to a sharp drop in performance: for example, \name{}, which attains \finalcorrect correctness with the assembly baseline, fails to produce any compilable code without it. Similar degradation occurs for other models such as llm-compiler-13b and Claude models. These results indicate that, at least in their current state, LLMs can act as superoptimizers building on compiler outputs, but cannot replace compilers themselves.

\paragraph{Case Studies.} We illustrate in Appendix~\ref{subsec:app:case} two representative examples where an LLM discovers optimizations beyond the reach of a state-of-the-art compiler. The first example is about instruction selection. The second example involves loop restructuring, along with function call optimization by replacing checked function variants with simpler equivalents.

\paragraph{Limitations and Potential Directions.}
Our evaluation relies on test-based validation, as in prior work~\citep{shypula2023learning,du2024mercury}, since existing assembly equivalence checkers cannot handle the complex control flow in our dataset~\citep{schkufza2013stochastic}. We mitigate this limitation with 96.2\% line coverage, which reduces the risk of undetected errors. More broadly, better verification tools for assembly remain an important direction, especially given recent progress on formal verification for GPU kernels~\citep{dubey2025equivalence}. Another promising direction is extending superoptimization beyond x86-64 to targets such as ARM and RISC-V. Finally, although our benchmark does not replace full systems benchmark suites, it is substantially more realistic than prior superoptimization benchmarks and provides a foundation for optimizing hot functions in real applications.
\section{Conclusion}
\label{sec:conclusion}

We investigated whether LLMs can act as superoptimizers, generating assembly programs that outperform code already optimized by industry-standard compilers. To support this study, we introduced the first large-scale benchmark of \numdataset assembly programs. Evaluating \numllmother{} models revealed the difficulty of the task, with most failing to achieve meaningful gains. By fine-tuning with reinforcement learning, our model \name{} improved both correctness and performance, reaching \finalcorrect test pass rate and an average speedup of \finalspeedup{} over gcc -O3. We also show that Best-of-N sampling and iterative refinement can bring additional improvement. These results demonstrate, for the first time, that LLMs can be applied as superoptimizers for assembly code, opening new opportunities for performance optimization beyond compiler heuristics.

\section*{Acknowledgments}
This work was supported in part by a Google Research Award.

\bibliography{colm2026_conference}

@inproceedings{pancompiler,
  title={Compiler-R1: Towards Agentic Compiler Auto-tuning with Reinforcement Learning},
  author={Pan, Haolin and Lin, Hongyu and Luo, Haoran and Liu, Yang and Yao, Kaichun and Zhang, Libo and Xing, Mingjie and Wu, Yanjun},
  booktitle={The Thirty-ninth Annual Conference on Neural Information Processing Systems},
  year={2025}
}

@inproceedings{lu2010contextual,
  title={Contextual multi-armed bandits},
  author={Lu, Tyler and P{\'a}l, D{\'a}vid and P{\'a}l, Martin},
  booktitle={Proceedings of the Thirteenth international conference on Artificial Intelligence and Statistics},
  pages={485--492},
  year={2010},
  organization={JMLR Workshop and Conference Proceedings}
}

@book{warren2013hacker,
  title={Hacker's delight},
  author={Warren, Henry S},
  year={2013},
  publisher={Pearson Education}
}

@article{sheng2024hybridflow,
  title={Hybridflow: A flexible and efficient rlhf framework},
  author={Sheng, Guangming and Zhang, Chi and Ye, Zilingfeng and Wu, Xibin and Zhang, Wang and Zhang, Ru and Peng, Yanghua and Lin, Haibin and Wu, Chuan},
  journal={arXiv preprint arXiv:2409.19256},
  year={2024}
}

@article{wolf1991loop,
  title={A loop transformation theory and an algorithm to maximize parallelism},
  author={Wolf, Michael E and Lam, Monica S},
  journal={IEEE transactions on parallel and distributed systems},
  volume={2},
  number={4},
  pages={452--471},
  year={1991}
}

@article{austin2021program,
  title={Program synthesis with large language models},
  author={Austin, Jacob and Odena, Augustus and Nye, Maxwell and Bosma, Maarten and Michalewski, Henryk and Dohan, David and Jiang, Ellen and Cai, Carrie and Terry, Michael and Le, Quoc and others},
  journal={arXiv preprint arXiv:2108.07732},
  year={2021}
}

@article{xia2023universal,
  title={Universal fuzzing via large language models},
  author={Xia, Chunqiu Steven and Paltenghi, Matteo and Le Tian, Jia and Pradel, Michael and Zhang, Lingming},
  journal={CoRR},
  year={2023}
}

@inproceedings{deng2024large,
  title={Large language models are edge-case generators: Crafting unusual programs for fuzzing deep learning libraries},
  author={Deng, Yinlin and Xia, Chunqiu Steven and Yang, Chenyuan and Zhang, Shizhuo Dylan and Yang, Shujing and Zhang, Lingming},
  booktitle={Proceedings of the 46th IEEE/ACM international conference on software engineering},
  pages={1--13},
  year={2024}
}

@inproceedings{xia2023automated,
  title={Automated program repair in the era of large pre-trained language models},
  author={Xia, Chunqiu Steven and Wei, Yuxiang and Zhang, Lingming},
  booktitle={2023 IEEE/ACM 45th International Conference on Software Engineering (ICSE)},
  pages={1482--1494},
  year={2023},
  organization={IEEE}
}

@inproceedings{theodoridis2022finding,
  title={Finding missed optimizations through the lens of dead code elimination},
  author={Theodoridis, Theodoros and Rigger, Manuel and Su, Zhendong},
  booktitle={Proceedings of the 27th ACM International Conference on Architectural Support for Programming Languages and Operating Systems},
  pages={697--709},
  year={2022}
}

@inproceedings{koenig2021adaptive,
  title={Adaptive restarts for stochastic synthesis},
  author={Koenig, Jason R and Padon, Oded and Aiken, Alex},
  booktitle={Proceedings of the 42nd ACM SIGPLAN International Conference on Programming Language Design and Implementation},
  pages={696--709},
  year={2021}
}

@software{artifact,
author = {Koenig, Jason R. and Padon, Oded and Aiken, Alex},
title = {Replication Package for Article: Adaptive Restarts for Stochastic Synthesis},
year = {2021},
publisher = {Association for Computing Machinery},
url = {https://doi.org/10.1145/3410298},
}

@inproceedings{shypula2023learning,
    title={Learning Performance-Improving Code Edits},
    author={Alexander Shypula and Aman Madaan and Yimeng Zeng and Uri Alon and Jacob R. Gardner and Yiming Yang and Milad Hashemi and Graham Neubig and Parthasarathy Ranganathan and Osbert Bastani and Amir Yazdanbakhsh},
    booktitle={The Twelfth International Conference on Learning Representations},
    year={2024},
    url={https://openreview.net/forum?id=ix7rLVHXyY}
}

@article{dubey2025equivalence,
  title={Equivalence Checking of ML GPU Kernels},
  author={Dubey, Kshitij and Driscoll, Benjamin and Wei, Anjiang and Kayal, Neeraj and Sharma, Rahul and Aiken, Alex},
  journal={arXiv preprint arXiv:2511.12638},
  year={2025}
}

@article{cummins2024meta,
  title={Meta large language model compiler: Foundation models of compiler optimization},
  author={Cummins, Chris and Seeker, Volker and Grubisic, Dejan and Roziere, Baptiste and Gehring, Jonas and Synnaeve, Gabriel and Leather, Hugh},
  journal={arXiv preprint arXiv:2407.02524},
  year={2024}
}

@inproceedings{deng2025compilerdream,
  title={Compilerdream: Learning a compiler world model for general code optimization},
  author={Deng, Chaoyi and Wu, Jialong and Feng, Ningya and Wang, Jianmin and Long, Mingsheng},
  booktitle={Proceedings of the 31st ACM SIGKDD Conference on Knowledge Discovery and Data Mining V. 2},
  pages={486--497},
  year={2025}
}

@article{ouyang2025kernelbench,
  title={KernelBench: Can LLMs Write Efficient GPU Kernels?},
  author={Ouyang, Anne and Guo, Simon and Arora, Simran and Zhang, Alex L and Hu, William and R{\'e}, Christopher and Mirhoseini, Azalia},
  journal={arXiv preprint arXiv:2502.10517},
  year={2025}
}

@inproceedings{zhao2024felix,
  title={Felix: Optimizing tensor programs with gradient descent},
  author={Zhao, Yifan and Sharif, Hashim and Adve, Vikram and Misailovic, Sasa},
  booktitle={Proceedings of the 29th ACM International Conference on Architectural Support for Programming Languages and Operating Systems, Volume 3},
  pages={367--381},
  year={2024}
}

@article{wei2025swe,
  title={Swe-rl: Advancing llm reasoning via reinforcement learning on open software evolution},
  author={Wei, Yuxiang and Duchenne, Olivier and Copet, Jade and Carbonneaux, Quentin and Zhang, Lingming and Fried, Daniel and Synnaeve, Gabriel and Singh, Rishabh and Wang, Sida I},
  journal={arXiv preprint arXiv:2502.18449},
  year={2025}
}

@article{le2022coderl,
  title={Coderl: Mastering code generation through pretrained models and deep reinforcement learning},
  author={Le, Hung and Wang, Yue and Gotmare, Akhilesh Deepak and Savarese, Silvio and Hoi, Steven Chu Hong},
  journal={Advances in Neural Information Processing Systems},
  volume={35},
  pages={21314--21328},
  year={2022}
}

@article{schkufza2013stochastic,
  title={Stochastic superoptimization},
  author={Schkufza, Eric and Sharma, Rahul and Aiken, Alex},
  journal={ACM SIGARCH Computer Architecture News},
  volume={41},
  number={1},
  pages={305--316},
  year={2013},
  publisher={ACM New York, NY, USA}
}

@article{puri2021codenet,
  title={Codenet: A large-scale ai for code dataset for learning a diversity of coding tasks},
  author={Puri, Ruchir and Kung, David S and Janssen, Geert and Zhang, Wei and Domeniconi, Giacomo and Zolotov, Vladimir and Dolby, Julian and Chen, Jie and Choudhury, Mihir and Decker, Lindsey and others},
  journal={arXiv preprint arXiv:2105.12655},
  year={2021}
}

@article{bunel2016learning,
  title={Learning to superoptimize programs},
  author={Bunel, Rudy and Desmaison, Alban and Kumar, M Pawan and Torr, Philip HS and Kohli, Pushmeet},
  journal={arXiv preprint arXiv:1611.01787},
  year={2016}
}

@article{sharma2015conditionally,
  title={Conditionally correct superoptimization},
  author={Sharma, Rahul and Schkufza, Eric and Churchill, Berkeley and Aiken, Alex},
  journal={ACM SIGPLAN Notices},
  volume={50},
  number={10},
  pages={147--162},
  year={2015},
  publisher={ACM New York, NY, USA}
}

@article{wu2024mirage,
  title={Mirage: A Multi-Level Superoptimizer for Tensor Programs},
  author={Wu, Mengdi and Cheng, Xinhao and Liu, Shengyu and Shi, Chunan and Ji, Jianan and Ao, Kit and Velliengiri, Praveen and Miao, Xupeng and Padon, Oded and Jia, Zhihao},
  journal={arXiv preprint arXiv:2405.05751},
  year={2024}
}

@article{churchill2017sound,
  title={Sound loop superoptimization for google native client},
  author={Churchill, Berkeley and Sharma, Rahul and Bastien, JF and Aiken, Alex},
  journal={ACM SIGPLAN Notices},
  volume={52},
  number={4},
  pages={313--326},
  year={2017},
  publisher={ACM New York, NY, USA}
}

@inproceedings{phothilimthana2016scaling,
  title={Scaling up superoptimization},
  author={Phothilimthana, Phitchaya Mangpo and Thakur, Aditya and Bodik, Rastislav and Dhurjati, Dinakar},
  booktitle={Proceedings of the Twenty-First International Conference on Architectural Support for Programming Languages and Operating Systems},
  pages={297--310},
  year={2016}
}

@article{liu2023rltf,
  title={Rltf: Reinforcement learning from unit test feedback},
  author={Liu, Jiate and Zhu, Yiqin and Xiao, Kaiwen and Fu, Qiang and Han, Xiao and Yang, Wei and Ye, Deheng},
  journal={arXiv preprint arXiv:2307.04349},
  year={2023}
}

@misc{hyperfine,
  title={Hyperfine},
  year={2025},
  url={https://github.com/sharkdp/hyperfine},
}

@article{schulman2017proximal,
  title={Proximal policy optimization algorithms},
  author={Schulman, John and Wolski, Filip and Dhariwal, Prafulla and Radford, Alec and Klimov, Oleg},
  journal={arXiv preprint arXiv:1707.06347},
  year={2017}
}

@article{shojaee2023execution,
  title={Execution-based code generation using deep reinforcement learning},
  author={Shojaee, Parshin and Jain, Aneesh and Tipirneni, Sindhu and Reddy, Chandan K},
  journal={arXiv preprint arXiv:2301.13816},
  year={2023}
}

@article{xia2024agentless,
  title={Agentless: Demystifying llm-based software engineering agents},
  author={Xia, Chunqiu Steven and Deng, Yinlin and Dunn, Soren and Zhang, Lingming},
  journal={arXiv preprint arXiv:2407.01489},
  year={2024}
}

@article{dou2024stepcoder,
  title={Stepcoder: Improve code generation with reinforcement learning from compiler feedback},
  author={Dou, Shihan and Liu, Yan and Jia, Haoxiang and Xiong, Limao and Zhou, Enyu and Shen, Wei and Shan, Junjie and Huang, Caishuang and Wang, Xiao and Fan, Xiaoran and others},
  journal={arXiv preprint arXiv:2402.01391},
  year={2024}
}

@inproceedings{liang2023learning,
  title={Learning compiler pass orders using coreset and normalized value prediction},
  author={Liang, Youwei and Stone, Kevin and Shameli, Ali and Cummins, Chris and Elhoushi, Mostafa and Guo, Jiadong and Steiner, Benoit and Yang, Xiaomeng and Xie, Pengtao and Leather, Hugh James and others},
  booktitle={International Conference on Machine Learning},
  pages={20746--20762},
  year={2023},
  organization={PMLR}
}

@article{haj2020autophase,
  title={Autophase: Juggling hls phase orderings in random forests with deep reinforcement learning},
  author={Haj-Ali, Ameer and Huang, Qijing Jenny and Xiang, John and Moses, William and Asanovic, Krste and Wawrzynek, John and Stoica, Ion},
  journal={Proceedings of machine learning and systems},
  volume={2},
  pages={70--81},
  year={2020}
}

@article{agrawal2025gepa,
  title={Gepa: Reflective prompt evolution can outperform reinforcement learning},
  author={Agrawal, Lakshya A and Tan, Shangyin and Soylu, Dilara and Ziems, Noah and Khare, Rishi and Opsahl-Ong, Krista and Singhvi, Arnav and Shandilya, Herumb and Ryan, Michael J and Jiang, Meng and others},
  journal={arXiv preprint arXiv:2507.19457},
  year={2025}
}

@article{hu2022lora,
  title={Lora: Low-rank adaptation of large language models.},
  author={Hu, Edward J and Shen, Yelong and Wallis, Phillip and Allen-Zhu, Zeyuan and Li, Yuanzhi and Wang, Shean and Wang, Lu and Chen, Weizhu and others},
  journal={ICLR},
  volume={1},
  number={2},
  pages={3},
  year={2022}
}

@article{shao2022tensor,
  title={Tensor program optimization with probabilistic programs},
  author={Shao, Junru and Zhou, Xiyou and Feng, Siyuan and Hou, Bohan and Lai, Ruihang and Jin, Hongyi and Lin, Wuwei and Masuda, Masahiro and Yu, Cody Hao and Chen, Tianqi},
  journal={Advances in Neural Information Processing Systems},
  volume={35},
  pages={35783--35796},
  year={2022}
}

@inproceedings{zheng2020ansor,
  title={Ansor: Generating $\{$High-Performance$\}$ tensor programs for deep learning},
  author={Zheng, Lianmin and Jia, Chengfan and Sun, Minmin and Wu, Zhao and Yu, Cody Hao and Haj-Ali, Ameer and Wang, Yida and Yang, Jun and Zhuo, Danyang and Sen, Koushik and others},
  booktitle={14th USENIX symposium on operating systems design and implementation (OSDI 20)},
  pages={863--879},
  year={2020}
}

@inproceedings{zheng2022amos,
  title={AMOS: enabling automatic mapping for tensor computations on spatial accelerators with hardware abstraction},
  author={Zheng, Size and Chen, Renze and Wei, Anjiang and Jin, Yicheng and Han, Qin and Lu, Liqiang and Wu, Bingyang and Li, Xiuhong and Yan, Shengen and Liang, Yun},
  booktitle={Proceedings of the 49th Annual International Symposium on Computer Architecture},
  pages={874--887},
  year={2022}
}

@article{chen2018learning,
  title={Learning to optimize tensor programs},
  author={Chen, Tianqi and Zheng, Lianmin and Yan, Eddie and Jiang, Ziheng and Moreau, Thierry and Ceze, Luis and Guestrin, Carlos and Krishnamurthy, Arvind},
  journal={Advances in Neural Information Processing Systems},
  volume={31},
  year={2018}
}

@article{liu2024learning,
  title={Learning code preference via synthetic evolution},
  author={Liu, Jiawei and Nguyen, Thanh and Shang, Mingyue and Ding, Hantian and Li, Xiaopeng and Yu, Yu and Kumar, Varun and Wang, Zijian},
  journal={arXiv preprint arXiv:2410.03837},
  year={2024}
}

@article{wei2025astra,
  title={Astra: A Multi-Agent System for GPU Kernel Performance Optimization},
  author={Wei, Anjiang and Sun, Tianran and Seenichamy, Yogesh and Song, Hang and Ouyang, Anne and Mirhoseini, Azalia and Wang, Ke and Aiken, Alex},
  journal={arXiv preprint arXiv:2509.07506},
  year={2025}
}

@article{team2023gemini,
  title={Gemini: a family of highly capable multimodal models},
  author={Team, Gemini and Anil, Rohan and Borgeaud, Sebastian and Alayrac, Jean-Baptiste and Yu, Jiahui and Soricut, Radu and Schalkwyk, Johan and Dai, Andrew M and Hauth, Anja and Millican, Katie and others},
  journal={arXiv preprint arXiv:2312.11805},
  year={2023}
}

@article{achiam2023gpt,
  title={Gpt-4 technical report},
  author={Achiam, Josh and Adler, Steven and Agarwal, Sandhini and Ahmad, Lama and Akkaya, Ilge and Aleman, Florencia Leoni and Almeida, Diogo and Altenschmidt, Janko and Altman, Sam and Anadkat, Shyamal and others},
  journal={arXiv preprint arXiv:2303.08774},
  year={2023}
}

@article{shao2024deepseekmath,
  title={Deepseekmath: Pushing the limits of mathematical reasoning in open language models},
  author={Shao, Zhihong and Wang, Peiyi and Zhu, Qihao and Xu, Runxin and Song, Junxiao and Bi, Xiao and Zhang, Haowei and Zhang, Mingchuan and Li, YK and Wu, Y and others},
  journal={arXiv preprint arXiv:2402.03300},
  year={2024}
}

@article{grubisic2024compiler,
  title={Compiler generated feedback for large language models},
  author={Grubisic, Dejan and Cummins, Chris and Seeker, Volker and Leather, Hugh},
  journal={arXiv preprint arXiv:2403.14714},
  year={2024}
}

@article{yang2024exploring,
  title={Exploring and unleashing the power of large language models in automated code translation},
  author={Yang, Zhen and Liu, Fang and Yu, Zhongxing and Keung, Jacky Wai and Li, Jia and Liu, Shuo and Hong, Yifan and Ma, Xiaoxue and Jin, Zhi and Li, Ge},
  journal={Proceedings of the ACM on Software Engineering},
  volume={1},
  number={FSE},
  pages={1585--1608},
  year={2024},
  publisher={ACM New York, NY, USA}
}

@article{bhatia2024verified,
  title={Verified code transpilation with LLMs},
  author={Bhatia, Sahil and Qiu, Jie and Hasabnis, Niranjan and Seshia, Sanjit and Cheung, Alvin},
  journal={Advances in Neural Information Processing Systems},
  volume={37},
  pages={41394--41424},
  year={2024}
}

@inproceedings{taneja2025llm,
  title={Llm-vectorizer: Llm-based verified loop vectorizer},
  author={Taneja, Jubi and Laird, Avery and Yan, Cong and Musuvathi, Madan and Lahiri, Shuvendu K},
  booktitle={Proceedings of the 23rd ACM/IEEE International Symposium on Code Generation and Optimization},
  pages={137--149},
  year={2025}
}

@article{ehrlich2025codemonkeys,
  title={Codemonkeys: Scaling test-time compute for software engineering},
  author={Ehrlich, Ryan and Brown, Bradley and Juravsky, Jordan and Clark, Ronald and R{\'e}, Christopher and Mirhoseini, Azalia},
  journal={arXiv preprint arXiv:2501.14723},
  year={2025}
}

@inproceedings{cummins2025llm,
  title={LLM Compiler: Foundation Language Models for Compiler Optimization},
  author={Cummins, Chris and Seeker, Volker and Grubisic, Dejan and Roziere, Baptiste and Gehring, Jonas and Synnaeve, Gabriel and Leather, Hugh},
  booktitle={Proceedings of the 34th ACM SIGPLAN International Conference on Compiler Construction},
  pages={141--153},
  year={2025}
}

@article{du2024mercury,
  title={Mercury: A code efficiency benchmark for code large language models},
  author={Du, Mingzhe and Luu, Anh Tuan and Ji, Bin and Liu, Qian and Ng, See-Kiong},
  journal={arXiv preprint arXiv:2402.07844},
  year={2024}
}

@article{li2025cuda,
  title={Cuda-l1: Improving cuda optimization via contrastive reinforcement learning},
  author={Li, Xiaoya and Sun, Xiaofei and Wang, Albert and Li, Jiwei and Shum, Chris},
  journal={arXiv preprint arXiv:2507.14111},
  year={2025}
}

@article{baronio2025kevin,
  title={Kevin: Multi-turn rl for generating cuda kernels},
  author={Baronio, Carlo and Marsella, Pietro and Pan, Ben and Guo, Simon and Alberti, Silas},
  journal={arXiv preprint arXiv:2507.11948},
  year={2025}
}

@article{liu2024evaluating,
  title={Evaluating language models for efficient code generation},
  author={Liu, Jiawei and Xie, Songrun and Wang, Junhao and Wei, Yuxiang and Ding, Yifeng and Zhang, Lingming},
  journal={arXiv preprint arXiv:2408.06450},
  year={2024}
}

@article{huang2024effilearner,
  title={Effilearner: Enhancing efficiency of generated code via self-optimization},
  author={Huang, Dong and Dai, Jianbo and Weng, Han and Wu, Puzhen and Qing, Yuhao and Cui, Heming and Guo, Zhijiang and Zhang, Jie},
  journal={Advances in Neural Information Processing Systems},
  volume={37},
  pages={84482--84522},
  year={2024}
}

@article{tehranijamsaz2024coderosetta,
  title={CodeRosetta: Pushing the Boundaries of Unsupervised Code Translation for Parallel Programming},
  author={TehraniJamsaz, Ali and Bhattacharjee, Arijit and Chen, Le and Ahmed, Nesreen K and Yazdanbakhsh, Amir and Jannesari, Ali},
  journal={arXiv preprint arXiv:2410.20527},
  year={2024}
}

@article{yang2024swe,
  title={Swe-agent: Agent-computer interfaces enable automated software engineering},
  author={Yang, John and Jimenez, Carlos and Wettig, Alexander and Lieret, Kilian and Yao, Shunyu and Narasimhan, Karthik and Press, Ofir},
  journal={Advances in Neural Information Processing Systems},
  volume={37},
  pages={50528--50652},
  year={2024}
}

@inproceedings{wei2025codearc,
title={Code{ARC}: Benchmarking Reasoning Capabilities of {LLM} Agents for Inductive Program Synthesis},
author={Anjiang Wei and Tarun Suresh and Jiannan Cao and Naveen Kannan and Yuheng Wu and Kai Yan and Thiago S. F. X. Teixeira and Ke Wang and Alex Aiken},
booktitle={Second Conference on Language Modeling},
year={2025},
url={https://openreview.net/forum?id=Q5pVZCrrKr}
}

@inproceedings{wei2025improving,
title={Improving Parallel Program Performance with {LLM} Optimizers via Agent-System Interfaces},
author={Anjiang Wei and Allen Nie and Thiago S. F. X. Teixeira and Rohan Yadav and Wonchan Lee and Ke Wang and Alex Aiken},
booktitle={Forty-second International Conference on Machine Learning},
year={2025},
url={https://openreview.net/forum?id=3h80HyStMH}
}

@article{jimenez2023swe,
  title={Swe-bench: Can language models resolve real-world github issues?},
  author={Jimenez, Carlos E and Yang, John and Wettig, Alexander and Yao, Shunyu and Pei, Kexin and Press, Ofir and Narasimhan, Karthik},
  journal={arXiv preprint arXiv:2310.06770},
  year={2023}
}

@inproceedings{xia2022less,
  title={Less training, more repairing please: revisiting automated program repair via zero-shot learning},
  author={Xia, Chunqiu Steven and Zhang, Lingming},
  booktitle={Proceedings of the 30th ACM Joint European Software Engineering Conference and Symposium on the Foundations of Software Engineering},
  pages={959--971},
  year={2022}
}

@article{li2022competition,
  title={Competition-level code generation with alphacode},
  author={Li, Yujia and Choi, David and Chung, Junyoung and Kushman, Nate and Schrittwieser, Julian and Leblond, R{\'e}mi and Eccles, Tom and Keeling, James and Gimeno, Felix and Dal Lago, Agustin and others},
  journal={Science},
  volume={378},
  number={6624},
  pages={1092--1097},
  year={2022},
  publisher={American Association for the Advancement of Science}
}

@article{chen2021evaluating,
  title={Evaluating large language models trained on code},
  author={Chen, Mark and Tworek, Jerry and Jun, Heewoo and Yuan, Qiming and Pinto, Henrique Ponde De Oliveira and Kaplan, Jared and Edwards, Harri and Burda, Yuri and Joseph, Nicholas and Brockman, Greg and others},
  journal={arXiv preprint arXiv:2107.03374},
  year={2021}
}

@article{hui2024qwen2,
  title={Qwen2. 5-coder technical report},
  author={Hui, Binyuan and Yang, Jian and Cui, Zeyu and Yang, Jiaxi and Liu, Dayiheng and Zhang, Lei and Liu, Tianyu and Zhang, Jiajun and Yu, Bowen and Lu, Keming and others},
  journal={arXiv preprint arXiv:2409.12186},
  year={2024}
}

@article{roziere2023code,
  title={Code llama: Open foundation models for code},
  author={Roziere, Baptiste and Gehring, Jonas and Gloeckle, Fabian and Sootla, Sten and Gat, Itai and Tan, Xiaoqing Ellen and Adi, Yossi and Liu, Jingyu and Sauvestre, Romain and Remez, Tal and others},
  journal={arXiv preprint arXiv:2308.12950},
  year={2023}
}

@article{touvron2023llama,
  title={Llama: Open and efficient foundation language models},
  author={Touvron, Hugo and Lavril, Thibaut and Izacard, Gautier and Martinet, Xavier and Lachaux, Marie-Anne and Lacroix, Timoth{\'e}e and Rozi{\`e}re, Baptiste and Goyal, Naman and Hambro, Eric and Azhar, Faisal and others},
  journal={arXiv preprint arXiv:2302.13971},
  year={2023}
}

@article{guo2025deepseek,
  title={Deepseek-r1: Incentivizing reasoning capability in llms via reinforcement learning},
  author={Guo, Daya and Yang, Dejian and Zhang, Haowei and Song, Junxiao and Zhang, Ruoyu and Xu, Runxin and Zhu, Qihao and Ma, Shirong and Wang, Peiyi and Bi, Xiao and others},
  journal={arXiv preprint arXiv:2501.12948},
  year={2025}
}

@article{liu2024deepseek,
  title={Deepseek-v3 technical report},
  author={Liu, Aixin and Feng, Bei and Xue, Bing and Wang, Bingxuan and Wu, Bochao and Lu, Chengda and Zhao, Chenggang and Deng, Chengqi and Zhang, Chenyu and Ruan, Chong and others},
  journal={arXiv preprint arXiv:2412.19437},
  year={2024}
}

@article{guo2024deepseek,
  title={DeepSeek-Coder: When the Large Language Model Meets Programming--The Rise of Code Intelligence},
  author={Guo, Daya and Zhu, Qihao and Yang, Dejian and Xie, Zhenda and Dong, Kai and Zhang, Wentao and Chen, Guanting and Bi, Xiao and Wu, Yu and Li, YK and others},
  journal={arXiv preprint arXiv:2401.14196},
  year={2024}
}

@article{li2023starcoder,
  title={Starcoder: may the source be with you!},
  author={Li, Raymond and Allal, Loubna Ben and Zi, Yangtian and Muennighoff, Niklas and Kocetkov, Denis and Mou, Chenghao and Marone, Marc and Akiki, Christopher and Li, Jia and Chim, Jenny and others},
  journal={arXiv preprint arXiv:2305.06161},
  year={2023}
}

@article{xia2024top,
  title={Top leaderboard ranking= top coding proficiency, always? evoeval: Evolving coding benchmarks via llm},
  author={Xia, Chunqiu Steven and Deng, Yinlin and Zhang, Lingming},
  journal={arXiv preprint arXiv:2403.19114},
  year={2024}
}

@article{zhuo2024bigcodebench,
  title={Bigcodebench: Benchmarking code generation with diverse function calls and complex instructions},
  author={Zhuo, Terry Yue and Vu, Minh Chien and Chim, Jenny and Hu, Han and Yu, Wenhao and Widyasari, Ratnadira and Yusuf, Imam Nur Bani and Zhan, Haolan and He, Junda and Paul, Indraneil and others},
  journal={arXiv preprint arXiv:2406.15877},
  year={2024}
}

@article{li2024evocodebench,
  title={Evocodebench: An evolving code generation benchmark with domain-specific evaluations},
  author={Li, Jia and Li, Ge and Zhang, Xuanming and Zhao, Yunfei and Dong, Yihong and Jin, Zhi and Li, Binhua and Huang, Fei and Li, Yongbin},
  journal={Advances in Neural Information Processing Systems},
  volume={37},
  pages={57619--57641},
  year={2024}
}

@article{hendrycks2021measuring,
  title={Measuring coding challenge competence with apps},
  author={Hendrycks, Dan and Basart, Steven and Kadavath, Saurav and Mazeika, Mantas and Arora, Akul and Guo, Ethan and Burns, Collin and Puranik, Samir and He, Horace and Song, Dawn and others},
  journal={arXiv preprint arXiv:2105.09938},
  year={2021}
}

@inproceedings{evalplus,
  title = {Is Your Code Generated by Chat{GPT} Really Correct? Rigorous Evaluation of Large Language Models for Code Generation},
  author = {Liu, Jiawei and Xia, Chunqiu Steven and Wang, Yuyao and Zhang, Lingming},
  booktitle = {Advances in Neural Information Processing Systems},
  year = {2023}
}

@book{thomas1971catalogue,
  title={A Catalogue of Optimizing Transformations},
  author={Thomas J. Watson IBM Research Center and Allen, F.E. and Cocke, J.},
  url={https://books.google.com/books?id=oeXaZwEACAAJ},
  year={1971},
  publisher={IBM Thomas J. Watson Research Center}
}
\bibliographystyle{colm2026_conference}

\appendix

\clearpage

\setcounter{figure}{0}
\renewcommand{\thefigure}{A\arabic{figure}}
\setcounter{table}{0}
\renewcommand{\thetable}{A\arabic{table}}

\section{Appendix}
\label{sec:app}

\subsection{Dataset Statistics}
\label{subsec:datasetstats}

\begin{table}[H]
\centering
\small
\setlength{\tabcolsep}{6pt}
\begin{tabular}{lccc|cc}
\toprule
\multirow{2}{*}{\textbf{Split}} & \multirow{2}{*}{\textbf{\# Prog.}} & \multirow{2}{*}{\textbf{Avg. Tests}} & \multicolumn{2}{c}{\textbf{Avg. LOC}} \\
& & & \textbf{C} & \textbf{Assembly} \\
\midrule
Training & 7,872  & 8.86 & 22.3 & 130.3 \\
Evaluation   &  200  & 8.92 & 21.9 & 133.3 \\
\bottomrule
\end{tabular}
\caption{Dataset statistics across training and evaluation splits. LOC = lines of code.}
\label{tab:dataset}
\end{table}

\subsection{Experimental Setup Details}
\label{subsec:app:setup}

\paragraph{Compiler Speedup.}  We quantify compiler optimizations by comparing gcc -O0 with gcc -O3 on the evaluation dataset and observe a mean speedup of 2.65×. This confirms that performance improvements in our dataset are measurable. Building on this baseline, we investigate whether LLMs can further enhance performance beyond the 2.65× speedup provided by the compiler.

\paragraph{Training Configurations.}
Among all evaluated models (see \Cref{tab:main}), we select \basemodel for training due to its strong correctness results and substantial room for performance improvement, while intentionally avoiding compiler-specific foundation models to preserve generality. Training is performed on a single node with four A100 GPUs. Hyperparameter settings are in \Cref{tab:training}. We train with PPO using the verl RL library, with both the actor and critic initialized from Qwen2.5-Coder-7B-Instruct, using a clip ratio of 0.2, an actor learning rate of 1e-6, a critic learning rate of 1e-5, and gradient clipping of 1.0 (clipping by norm). Rollouts are generated using vLLM on 4 A100 GPUs with FSDP, with a training batch size of 16 and a PPO mini-batch size of 16. We run a single epoch over the 7,872 training problems, giving approximately 492 optimization updates each for the actor and critic. We also train a GRPO variant using the same settings, with a group size of 5 (5 responses sampled per prompt) and 16 groups per batch.

\begin{table}[H]
\centering
\small
\begin{tabular}{ll}
\toprule
\textbf{Component} & \textbf{Setting} \\
\midrule
Base model & \texttt{Qwen2.5-Coder-7B-Instruct} \\
Actor's learning rate & 1e-6 \\
Critic's learning rate & 1e-5 \\
Batch size & 16 \\
Epoch & 1 \\
Max prompt length & 2000 tokens \\
Max response length & 2000 tokens \\
Gradient checkpointing & Enabled (both actor and critic) \\
Rollout temperature & 0.5 \\
Hardware & 4× A100 GPUs \\
\bottomrule
\end{tabular}
\caption{Key training configurations for PPO fine-tuning.}
\label{tab:training}
\vspace{1em}
\end{table}

\subsection{Prompt Template}
\label{subsec:prompttemplate}

\begin{tcolorbox}[
  colback=blue!3,
  colframe=gray!60!black,
  title=Prompt Template,
  sharp corners,
  fontupper=\ttfamily\small
]
Given the following C code and assembly code, your task is to generate highly optimized x86-64 assembly code.\\

C Code:
\begin{verbatim}
<C code here>
\end{verbatim}

Assembly Code:
\begin{verbatim}
<baseline assembly code here produced by gcc -O3>
\end{verbatim}

Only output the optimized assembly code. Do not include any other text. Do not write any comments in the assembly code.  Wrap the assembly code in assembly tags.

Optimized Assembly Code:
\end{tcolorbox}

\begin{table}[H]
\centering
\small
\begin{tabular}{lcccc}
\toprule
\textbf{Model} & \textbf{Shots} & \textbf{Compile Pass (\%)} & \textbf{Test Pass (\%)} & \textbf{Avg. Speedup} \\
\midrule
claude-opus-4           & 0-shot & 90.0 & 51.5 & 1.43× \\
claude-opus-4           & 2-shot & 95.0 & 15.0 & 1.13× \\
claude-opus-4           & 4-shot & 95.0 & 12.5 & 1.11× \\
SuperCoder (PPO)        & 0-shot & 96.0 & 95.0 & 1.46× \\
SuperCoder (PPO)        & 2-shot & 94.0 & 90.5 & 1.59× \\
SuperCoder (PPO)        & 4-shot & 93.0 & 81.0 & 1.54× \\
Qwen2.5-Coder-7B (Base) & 0-shot & 77.9 & 61.4 & 1.10× \\
Qwen2.5-Coder-7B (Base) & 2-shot & 70.5 & 35.0 & 1.10× \\
Qwen2.5-Coder-7B (Base) & 4-shot & 80.5 & 30.5 & 1.06× \\
\bottomrule
\end{tabular}
\vspace{1em}
\caption{Comparison of 0-shot, 2-shot, and 4-shot prompting across different models.}
\label{tab:shots}
\end{table}

\begin{figure}[H]
  \centering
  \includegraphics[scale=0.4]{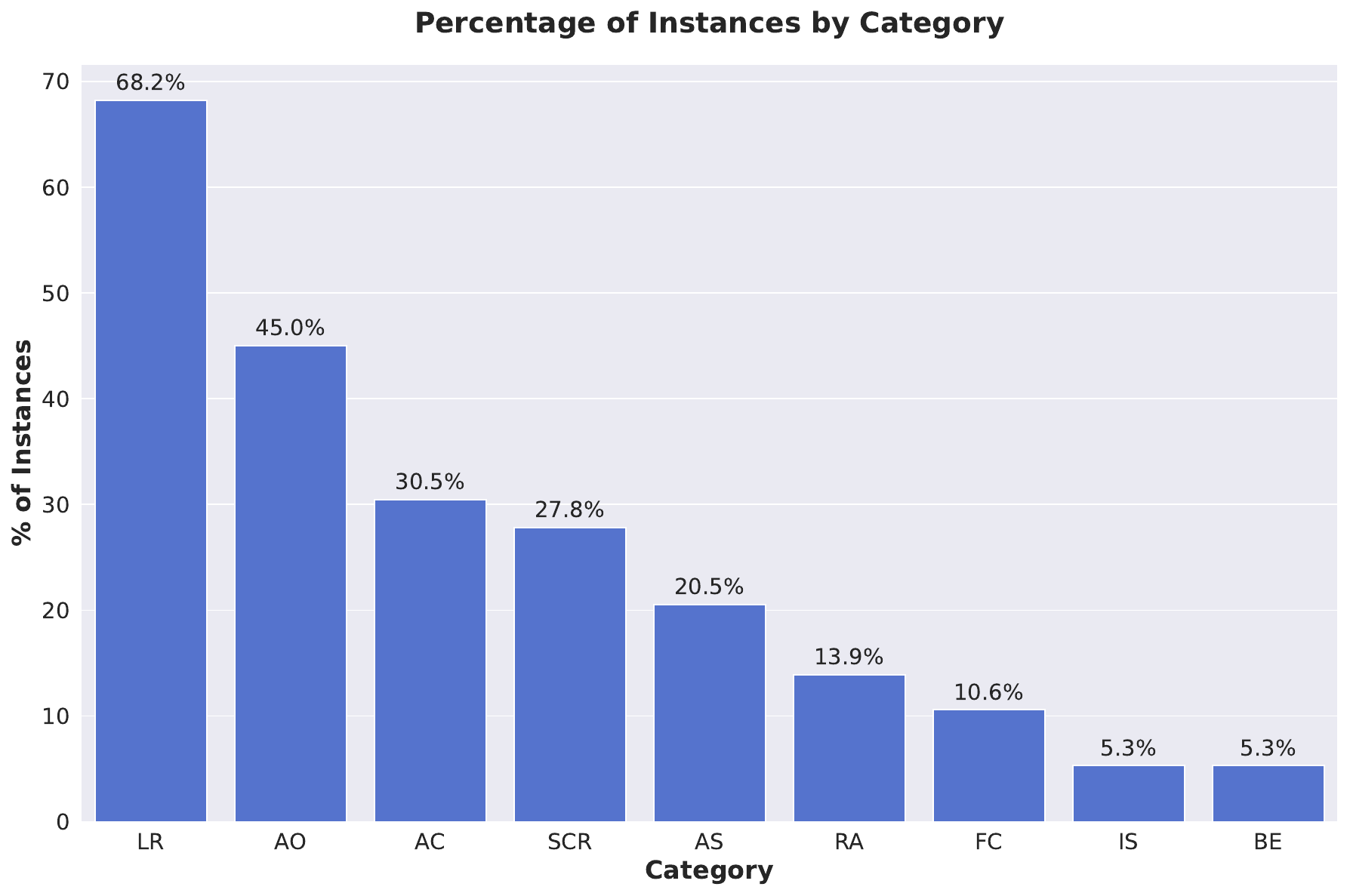}

  \caption{Categorization of the program transformations: Loop Restructuring (LR), Arithmetic Optimization (AO), Address Calculation (AC), Stack Canary Removal (SCR), Algorithmic Simplification (AS), Register Allocation (RA), Function Call Optimization (FC), Instruction Selection (IS), Branch Elimination (BE).}
  \label{fig:category}
\end{figure}

\subsection{Case Studies}
\label{subsec:app:case}

We show two case studies in \Cref{fig:casestudy} and \Cref{fig:casestudytwo}.

\begin{figure}[h]
  \centering
  \begin{minipage}[t]{0.3\textwidth}
  \centering
  \textbf{\fontsize{11}{13}\selectfont C Code}
  \vspace{0.5em}
  \begin{lstlisting}[style=cstyle]
int f(unsigned long x)
{
  int res = 0;
  while (x > 0)
  {
    res += x & 1;
    x >>= 1;
  }
  return res;
}
\end{lstlisting}
      
  \end{minipage}
  \hfill
  \begin{minipage}[t]{0.3\textwidth}
  \centering
  \textbf{\fontsize{11}{13}\selectfont GCC -O3 Output}
  \vspace{0.5em}
  \begin{lstlisting}[style=asmstyle]
  .L0:
  xorl %eax, %eax
  testq %rdi, %rdi
  je .L2
  .L1:
  movq %rdi, %rdx
  andl $0x1, %edx
  addq %rdx, %rax
  shrq $0x1, %rdi
  jne .L1
  retq
  .L2:
  retq
  \end{lstlisting}
  \end{minipage}
  \hfill
  \begin{minipage}[t]{0.3\textwidth}
  \centering
  \textbf{\fontsize{11}{13}\selectfont Claude-Opus-4}
  \vspace{0.5em}
  \begin{lstlisting}[style=asmstyle]
  .L0:
  popcnt %rdi, %rax
  retq
  \end{lstlisting}
  \end{minipage}
\caption{Case study comparing the C code, baseline assembly produced by gcc -O3, and optimized assembly generated by Claude-Opus-4. The model successfully replaces the loop with the specialized hardware instruction \texttt{popcnt}, resulting in a significantly more concise implementation.}
\label{fig:casestudy}
\end{figure}

\begin{figure}[h]
  \centering
  \hfill
  \begin{minipage}[t]{0.48\textwidth}
  \centering
  \textbf{\fontsize{11}{13}\selectfont GCC -O3 Output}
  \vspace{0.5em}
  \begin{lstlisting}[style=asmstyle]
    ...
    testl   %eax, %eax
    jle     .L2                
    movl    %eax, %r8d
    jmp     .L5                 
    .p2align 4,,10
    .p2align 3
.L4:
    subl    $1, %r8d            
    je      .L2                 
.L5:
    movl    %ecx, %eax
    cltd
    idivl   %r8d                
    testl   %edx, %edx          
    jne     .L4                 
    movl    %r8d, %edx
    leaq    .LC1(%rip), %rsi
    movl    $1, %edi
    xorl    %eax, %eax
    call    __printf_chk@PLT
.L2:
    movq	8(%rsp), %rax
    subq	%fs:40, %rax
    jne	.L14
    xorl	%eax, %eax
    addq	$24, %rsp
    .cfi_remember_state
    .cfi_def_cfa_offset 8
    ret
    ...
  \end{lstlisting}
  \end{minipage}
  \hfill
  \begin{minipage}[t]{0.48\textwidth}
  \centering
  \textbf{\fontsize{11}{13}\selectfont Claude-Opus-4}
  \vspace{0.5em}
  \begin{lstlisting}[style=asmstyle]
    ...
    testl   %eax, %eax
    jle     .L2                 
.L3:                            
    movl    %ecx, %edx
    movl    %eax, %esi
    movl    %edx, %eax
    cltd
    idivl   %esi                
    testl   %edx, %edx          
    je      .L4                 
    decl    %esi                
    movl    %esi, %eax
    jg      .L3                 
.L2:
    xorl    %eax, %eax
    addq    $24, %rsp
    ret
.L4:
    leaq	.LC1(%rip), %rdi
    xorl	%eax, %eax
    call	printf@PLT
    xorl	%eax, %eax
    addq	$24, %rsp
    ret
    .size	main, .-main
  \end{lstlisting}
  \end{minipage}
\caption{Case study comparing the baseline assembly code snippet produced by gcc -O3 and the optimized assembly code snippet generated by Claude-Opus-4. Claude-Opus-4 eliminates the entry-path unconditional jump and alignment padding by fusing GCC’s two-block loop into a single, simpler control-flow structure, calls \texttt{printf@PLT} directly (a simpler function variant), and removes gcc’s stack-protector canary check.}
\label{fig:casestudytwo}
\end{figure}

\subsection{Categorization of Program Transformations}
\label{subsec:app:trans}

We categorize the differences between the code optimized by claude-opus-4 with best-of-8 sampling and the compiler-generated code into nine classes, each capturing a distinct kind of optimization or transformation: 1) Loop Restructuring (LR) covers changes to loop organization, including reordering, unrolling, or altered control flow; 2) Arithmetic Optimization (AO) includes simplifying arithmetic operations such as replacing divisions or multiplications with shifts, or using increment/decrement instructions where applicable; 3) Address Calculation (AC) optimization refers to simplifying memory address computations and offset arithmetic; 4) Stack Canary Removal (SCR) refers to eliminating stack protection checks and related instrumentation; 5) Algorithmic Simplification (AS) denotes replacing complex or custom logic with simpler algorithms or standard library routines such as memcmp, strcmp, or atoi; 6) Register Allocation (RA) reflects differences in how registers are assigned, reused, or spilled to memory; 7) Function Call (FC) optimization captures substituting heavyweight or checked function variants with simpler equivalents (e.g., \_\_isoc99\_scanf vs. scanf); 8) Instruction Selection (IS) captures the use of specialized CPU instructions (e.g., bsrq, popcnt, cmov) in place of longer generic instruction sequences; 9) Branch Elimination involves removing conditional branches by using conditional moves (cmov) or condition-setting instructions (setcc). In ~\Cref{fig:category}, we plot the percentage of instances in our evaluation dataset that have each of these categories. Loop Restructuring (LR) is the most frequent category, appearing in 68.2\% of instances, indicating that Claude often performs high-level changes to loop structure and control flow. Arithmetic Optimization (AO) and Address Calculation (AC) are also common, occurring in 45.0\% and 30.5\% of cases, respectively, while Stack Canary Removal (SCR) and Algorithmic Simplification (AS) appear in roughly a quarter of the dataset. Note that the total percentage exceeds 100\% because an instance may have multiple transformation types.

\end{document}